\documentclass[sigconf]{acmart}

\renewcommand\footnotetextcopyrightpermission[1]{} 
\usepackage{footmisc} 

\usepackage{booktabs} 
\usepackage{subfigure}
\usepackage{hhline}

\graphicspath{{images/}}

\setcopyright{rightsretained}

\begin{document}
\title{Deep Cross-Modal Audio-Visual Generation}


\author{Lele Chen\footnote[1]{}}
\affiliation{%
  \institution{Computer Science}
  \institution{University of Rochester}
}
\email{lchen63@cs.rochester.edu}

\author{Sudhanshu Srivastava}
\authornote{These authors contributed equally to this work.}
\affiliation{%
  \institution{Computer Science}
  \institution{University of Rochester}
}
\email{ssrivas6@cs.rochester.edu}

\author{Zhiyao Duan}
\affiliation{%
  \institution{Electrical and Computer Engineering}
  \institution{University of Rochester}
}
\email{zhiyao.duan@rochester.edu}

\author{Chenliang Xu}
\affiliation{%
  \institution{Computer Science}
  \institution{University of Rochester}
}
\email{chenliang.xu@rochester.edu}


\begin{abstract}
Cross-modal audio-visual perception has been a long-lasting topic in psychology and neurology, and various studies have discovered strong correlations in human perception of auditory and visual stimuli. Despite works in computational multimodal modeling, the problem of cross-modal audio-visual generation has not been systematically studied in the literature. In this paper, we make the first attempt to solve this cross-modal generation problem leveraging the power of deep generative adversarial training. Specifically, we use conditional generative adversarial networks to achieve cross-modal audio-visual generation of musical performances. We explore different encoding methods for audio and visual signals, and work on two scenarios: instrument-oriented generation and pose-oriented generation. Being the first to explore this new problem, we compose two new datasets with pairs of images and sounds of musical performances of different instruments. Our experiments using both classification and human evaluations demonstrate that our model has the ability to generate one modality, i.e., audio/visual, from the other modality, i.e., visual/audio, to a good extent. Our experiments on various design choices along with the datasets will facilitate future research in this new problem space. 
\end{abstract}





\maketitle

%
%
%
%

\begin{figure*}[ht!]
\centering
\includegraphics[width=\linewidth]{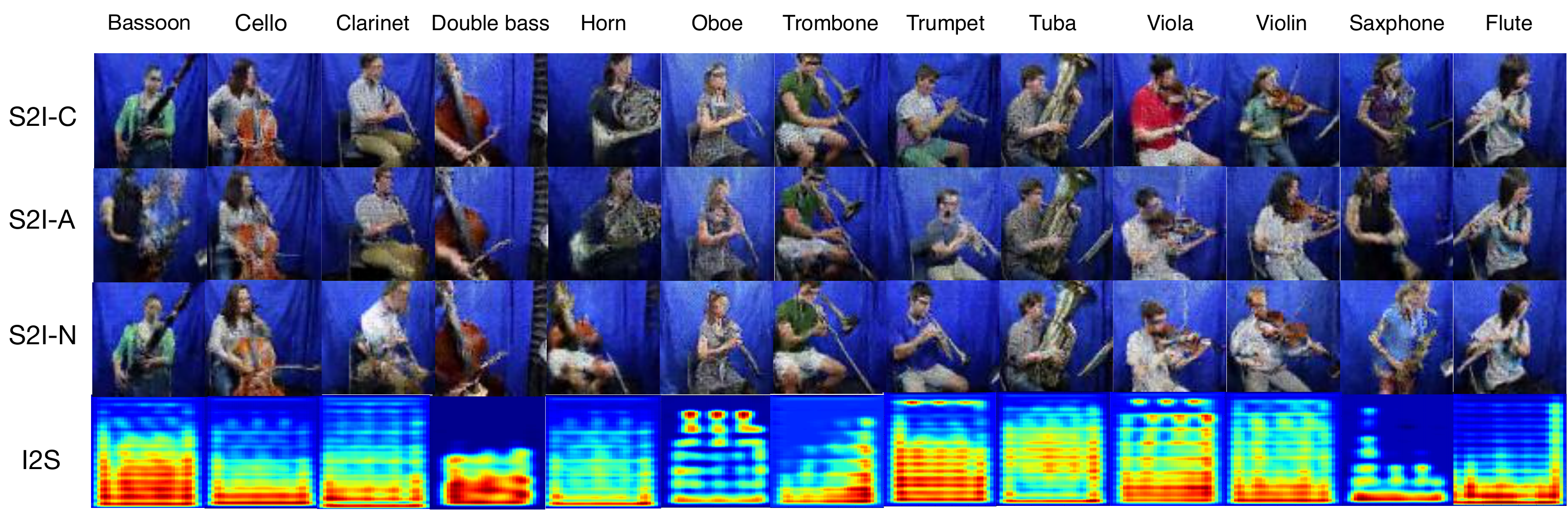}
\caption{Generated outputs using our cross-modal audio-visual generation models. Top three rows are musical performance images generated by our Sound-to-Image (S2I) networks from audio recordings. S2I-C is our main model. S2I-A and S2I-N are variations of our main model. Bottom row contains the log-mel spectrograms of generated audio of different instruments from musical performance images using our Image-to-Sound (I2S) network. Each column represents one instrument type. 
}
\label{fig:result_training_part}
\end{figure*}

\section{Introduction}
\label{sec:intro}


Cross-modal perception, or intersensory phenomenon, has been a long-lasting research topic in numerous disciplines such as psychology~\cite{DaRoRuNEURO1973, StTHESIS1998, VrGeJEP2000, ViKrWaCOG2006} , neurology~\cite{StMeBOOK1993}, and human-computer interaction~\cite{MiVaCaCHI1993, TaLiHoACMMM2015}, and recently gained attention in computer vision~\cite{OwIsMcCVPR2016}, audition~\cite{LiDiDuICASSP2017} and multimedia analysis~\cite{PeCoDoTPAMI2014, FeWaLiACMMM2014}. In this paper, we focus on the problem of cross-modal audio-visual generation. Our system is trained with pairs of visual and audio signals, which are typically contained in videos, and is able to generate one modality (visual/audio) given observations from the other modality (audio/visual). Fig.~\ref{fig:result_training_part} shows results generated by our system on a musical performance video dataset. 

Learning from multimodal input is challenging---despite the many works in cross-modal analysis, a large portion of the effort, e.g.,~\cite{RaCoCoACMMM2010, PeCoDoTPAMI2014, FeWaLiACMMM2014, WaYiWaARXIV2016}, has been focused on indexing and retrieval instead of generation. Although joint representations of multiple modalities and their correlations are explored, these methods only need to retrieve samples that exist in a database. They do not, for example, need to model the details of the samples, which is required in data generation. 
On the contrary, the generation task requires generating novel images and sounds that are \textit{unseen} or \textit{unheard}, and is of great interest to many applications, such as creating art works~\cite{IsZhZhARXIV2016, ZhDaARXIV2017} and zero-shot learning~\cite{ChChGoECCV2016}. It requires learning a complex generative function that produces meaningful outputs. In the case of cross-modality generation, this function has to map from one modality space to the other modality space, making the problem even more challenging and interesting.

Generative Adversarial Networks (GANs)~\cite{GoPoMiNIPS2014} has become an emerging topic in deep generative models. Inspired by Reed et al.'s work on generating images conditioned on text captions ~\cite{ReAkYaICML2016}, we design Conditional GANs for cross-modal audio-visual generation. Different from their work, we make the networks to handle intersensory generation---generate images conditioned on sounds and generate sounds conditioned on images. We explore two different tasks when generating images: instrument-oriented generation (see Fig.~\ref{fig:result_training_part}) and pose-oriented generation (see Fig.~\ref{fig:pose}), where the latter task is treated as fine-grained generation comparing to the former. 

Another key aspect to the success of cross-modal generation is being able to effectively encode and decode information contained in different modalities. For images, Convolutional Neural Networks (CNNs) are known to perform well in various tasks. Therefore, we train a CNN and use the fully connected layer before softmax as the image encoder and use several deconvolution layers as the decoder/generator. For sounds, we also use CNNs to encode and decode. The input to the networks, however, cannot be the raw waveforms. Instead, we first transform the time-domain signal into the time-frequency or time-quefrency domain. We explore five different transformations and find that the log-mel spectrogram gives the best result.


To explore this new problem space, we compose two datasets, e.g., Sub-URMP and INIS. The Sub-URMP dataset consists of paired images and sounds extracted from 107 single-instrument musical performance videos of 13 kinds of instruments in the University of Rochester Musical Performance (URMP) dataset~\cite{LiLiDiARXIV2016}. In total 17,555 images are extracted and each image is paired with a half-second long sound clip. The INIS dataset contains ImageNet~\cite{DeDoSoCVPR2009} images of five music instruments, e.g., drum, saxophone, piano, guitar and violin. We pair each image with a short sound clip of a solo performance of the corresponding instrument. We conduct experiments to evaluate the quality of our generated images and sound spectrograms using both classification and human evaluation. Our experiments demonstrate that our conditional GANs can, indeed, generate one modality (visual/audio) from the other modality (audio/visual) to a good extent at both the instrument-level and the pose-level. We also compare and evaluate various design choices in our experiments. 

The contributions are three-fold. First, to our best knowledge, we introduce the problem of cross-modal audio-visual generation and are the first to use GANs on intersensory generation. Second, we propose newnetwork structures and adversarial training strategies for cross-modal GANs. Third, we compose two datasets that will be released to facilitate future research in this new problem space. 

The paper is organized as follows. We discuss related work and background in Sec.~\ref{sec:related}. We introduce our network structure, training strategies and encoding methods in Sec.~\ref{sec:model}. We present our datasets in Sec.~\ref{sec:data} and experiments in Sec.~\ref{sec:exp}. Finally, we conclude our paper in Sec.~\ref{sec:con}.

\section{Related Work}
\label{sec:related}

Our work differs from the various works in cross-modal retrieval \cite{WaYiWaARXIV2016, RaCoCoACMMM2010, FeWaLiACMMM2014, PeCoDoTPAMI2014} as stated in Sec.~\ref{sec:intro}. In this section, we further distinguish our work from those in multimodal representation learning. Ngiam et al.~\cite{NgKhKiICML2011} learn a shared representation between audio-visual modalities by training a stacked multimodal autoencoder. Srivastava and Salakhutdinov~\cite{SrSaNIPS2012} propose a multimodal deep Boltzmann machine to learn a joint representation of images and their text tags. Kumar et al.~\cite{KuDhCoAAAI2014} learn an audio-visual bimodal compositional model using sparse coding. Our work differs from them by using the adversarial training framework that allows us to learn a much deeper representation for the generator. 

Adversarial training has recently received a significant amount of attention~\cite{MaShJaICLR2016, RaMeChICLR2015, ReAkYaICML2016, DeChSzNIPS2015, GoPoMiNIPS2014, SaGoZaNIPS2016, BeZiNIPSW2016}. It has been shown to be effective in various tasks, such as generating semantic segmentations~\cite{LuCoChARXIV2016, SoSpShARXIV2017}, improving object localization~\cite{BeZiNIPSW2016}, image-to-image translation~\cite{IsZhZhARXIV2016} and enhancing speech~\cite{PaBoSeARXIV2017}. We also use adversarial training but on a novel  problem of cross-modal audio-visual generation with music instruments and human poses that differs from other works. 


\begin{figure*}[t]
\centering
\includegraphics[width=1\linewidth]{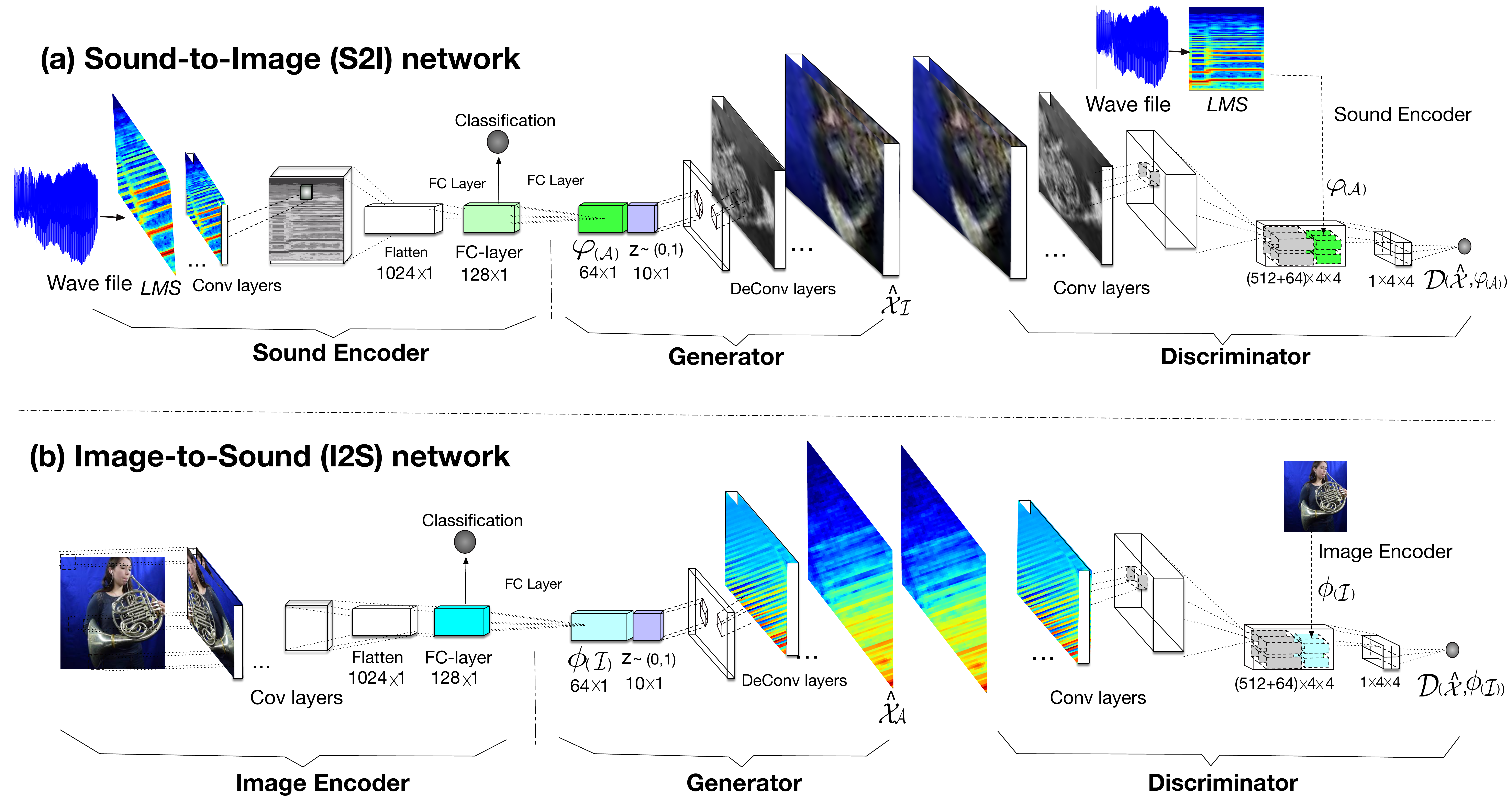}
\caption{The overall diagram of our model. This figure consists of an S2I GAN network (a) and an I2S GAN network (b). Each network contains an encoder, a generator and a discriminator respectively.} 
\label{fig:model}
\end{figure*}

\subsection{Background}
\label{sec:related:gan}

Generative Adversarial Networks (GANs) are introduced in the seminal work of Goodfellow et al.~\cite{GoPoMiNIPS2014}, and consist of a generator network $G$ and a discriminator network $D$. Given a distribution, $G$ is trained to generate samples that are resembled from this distribution, while $D$ is trained to distinguish whether the sample is genuine. They are trained in an adversarial fashion playing a min-max game against each other:
\begin{align}
\underset{G}{\text{min}}\, \underset{D}{\text{max}}\, V(D,G) = \,
&\mathbb{E}_{x \sim p_{data}(x)} [\log D(x)] + \\ \nonumber
&\mathbb{E}_{x \sim p_z(z)} [\log (1 - D(G(z)))]
\enspace,
\label{eq:gan}
\end{align}
where $p_{data}$ is the target data distribution and $z$ is drawn from a random noise  distribution $p_z$.

Conditional GANs~\cite{MiOsARXIV2014, DeChSzNIPS2015} are variants of GANs, where one is interested in directing the generation conditioned on some variables, e.g., labels in a dataset. It has the following form: 
\begin{align}
\underset{G}{\text{min}}\, \underset{D}{\text{max}}\, V(D,G) = \,
&\mathbb{E}_{x \sim p_{data}(x)} [\log D(x|y)] + \\ \nonumber
&\mathbb{E}_{x \sim p_z(z)} [\log (1 - D(G(z|y)))]
\enspace,
\end{align}
where the only difference from GANs is the introduction of $y$ that represents the condition variable. This condition is passed to both the generator and the discriminator networks. One particular example is \cite{ReAkYaICML2016}, where they use conditional GANs to generate images conditioned on text captions. The text captions are encoded through a recurrent neural network as in~\cite{ReAkLeCVPR2016}. In this paper, we use conditional GANs for cross-modal audio-visual generation.


\section{Cross-Modal Generation Model}
\label{sec:model}

The overall diagram of our model is shown in Fig.~\ref{fig:model}, where we have separate networks for Sound-to-Image (S2I) and Image-to-Sound (I2S). Each of them consists of three parts: an encoder network, a generator network, and a discriminator network. We describe the generator and discriminator networks in Sec.~\ref{sec:model:arch}, and their training strategies in Sec.~\ref{sec:model:train}. We present the encoder networks for sound and image in Sec.~\ref{sec:model:audioencode} and Sec.~\ref{sec:model:visualencode}, respectively. 

\subsection{Generator and Discriminator Networks}
\label{sec:model:arch}

\noindent \textbf{S2I Generator} \quad The S2I generator network is denoted as: $G_{S \mapsto I}: \mathbb{R}^{|\varphi(A)|} \times \mathbb{R}^Z \mapsto \mathbb{R}^{I}$. The sound encoding vector of size 128 is first compressed to a vector of size 64 via a fully connected layer followed by a leaky ReLU, which is denoted as $\varphi(A)$. Then it is concatenated with a random noise vector $z \in \mathbb{R}^Z$. The generator takes this concatenated vector and produces a synthetic image $\hat{x}_{I} \gets G_{S \mapsto I}(z,\varphi(A))$ of size 64x64x3. 

\noindent \textbf{S2I Discriminator} \quad The S2I discriminator network is denoted as: $D_{S \mapsto I}: \mathbb{R}^{I} \times \mathbb{R}^{|\varphi(A)|} \mapsto [0,1]$. It takes an image and a compressed sound encoding vector and produces a score for this pair being a genuine pair of image and sound. 

\noindent \textbf{I2S Generator} \quad Similarly, the I2S generator network is denoted as: $G_{I \mapsto S}: \mathbb{R}^{|\phi(I)|} \times \mathbb{R}^Z \mapsto \mathbb{R}^{A}$. The image encoding vector of size 128 is compressed to size 64 via a fully connected layer followed by a leaky ReLU, denoted as $\phi(I)$, and concatenated with a noise $z$. The generator takes it and do a forward pass to produce a synthetic sound spectrogram $\hat{x}_A \gets G_{I \mapsto S}(z,\phi(I))$ of size 128x34. 

\noindent \textbf{I2S Discriminator} \quad The I2S discriminator network is denoted as: $D_{I \mapsto S}: \mathbb{R}^A \times \mathbb{R}^{|\phi(I)|} \mapsto [0,1]$. It takes a sound spectrogram and a compressed image encoding vector and produces a score for this pair being a genuine pair of sound and image. 

Our implementation is based on the GAN-CLS by Reed et al.~\cite{ReAkYaICML2016}. We extend it to handle the challenges in operating sound spectrograms which have a rectangle size. For the I2S generator network, after getting a 32x32x128 feature map, we apply two successive deconvolution layers, where each has a kernel of size 4x4 with stride 2x1 and 1x1 zero-padding, and obtain a matrix of size 128x34. We apply the numpy resize function to get a matrix of size 128x44 for comparing with ground-truth spectrogram in evaluation. The I2S discriminator network takes sound spectrogram of size 128x34. To handle ground-truth spectrogram, we resize it from 128x44 to 128x34.
We apply two successive convolution layers, where each has a kernel of size 4x4 with stride 2x1 and 1x1 zero-padding. This results in a 32x32 square feature map. In practice, we have observed that adding more convolution layers in the I2S networks helps get better output in fewer epochs. We add two layers to the generator network and 12 layers to the discriminator network. 

\subsection{Adversarial Training Strategies}
\label{sec:model:train}

Without loss of generality, we assume that the training set contains pairs of images and sounds $\{(I_{i}^{j}, A_{i}^{j})\}$, where $I_{i}^{j}$ represents the $j$th image of the  $i$th instrument category in our dataset and $A_{i}^{j}$ represents the corresponding sound. Here, $i\in\{1,2,3\dots,13\}$ represents the index to one of the music instruments in our dataset, e.g., \textit{cello} or \textit{violin}. Notice that even images and sounds within the same music instrument category differ in terms of the player, pose, and music note. We use $\mathbf{I}_{-i}$ to represent the set of all images of instruments of all the categories except the $i$th category, and use $\mathbf{I}^{-j}_{i}$ to represent the set of all images in the $i$th instrument category except the $j$th image. The sound counterparts, $\mathbf{A}_{-i}$ and $\mathbf{A}^{-j}_{i}$, are defined likewise.

Based on the input, we define three kinds of discriminator outputs: $S_r$, $S_f$ and $S_w$. Here, $S_r$ is the score for a \textit{true} pair of image and sound that is contained in our training set, and $S_f$ is the score for the pair where one modality is \textit{generated} based on the other modality, and $S_w$ is the score for the wrong pair of image and sound. Wrong pairs are sampled from the training dataset. The generator network is trained to maximize:  
\begin{align}
\log(S_f)
\enspace,
\end{align}
and the discriminator is trained to maximize:
\begin{align}
\log(S_r) + (\log(1-S_w) + \log(1-S_f))/2
\enspace. 
\end{align}
Notice that by using different types of wrong pairs, we can eventually guide the generator in solving various tasks. 

\noindent \textbf{S2I Generation (Instrument-Oriented)} \quad 
We train a single S2I model over the entire dataset so that it can generate musical performance images of different instruments from different input sounds. In other words, the same model can generate an image of person-playing-violin from an unheard sound of violin, and can generate an image of person-playing-saxophone from an unheard sound of saxophone.
We apply the following training settings: 
\begin{align}
\hat{x}_I &\gets G_{S \mapsto I} (\varphi(A_{i}^{j}),z) \nonumber \\
S_f &= D_{S \mapsto I} (\hat{x}_I, \varphi(A_{i}^{j})) \nonumber \\
S_r &= D_{S \mapsto I} (I_{i}^{j}, \varphi(A_{i}^{j})) \nonumber \\
S_w &= D_{S \mapsto I} (\omega(\mathbf{I}_{-i}), \varphi(A_{i}^{j}))
\label{eq:s2i_instrument}
\enspace,
\end{align}
where $\hat{x}_I$ is the synthetic image of size 64x64x3, $z$ is the random noise vector and $\varphi(A_{i}^{j})$ is the compressed sound encoding. $\omega(\cdot)$ is a random sampler with a uniform distribution, and it samples images from the wrong instrument category to construct wrong pairs for calculating $S_w$. We use the sound-to-image network structure as in Fig.~\ref{fig:model} (a). 

\noindent \textbf{S2I Generation (Pose-Oriented)} \quad 
We train a set of S2I models with one for each music instrument category. Each model captures the relations between different human poses and input sounds of one instrument. For example, the model trained on violin image-sound pairs can generate a series of images of person-playing-violin with different hand movements according to different violin sounds. This is a fine-grained generation task compared to the previous instrument-oriented task. We apply the following training settings: 
\begin{align}
\hat{x}_I &\gets G_{S \mapsto I} (\varphi(A_{i}^{j}),z) \nonumber \\
S_f &= D_{S \mapsto I} (\hat{x}_I, \varphi(A_{i}^{j})) \nonumber \\
S_r &= D_{S \mapsto I} (I_{i}^{j}, \varphi(A_{i}^{j})) \nonumber \\
S_w &= D_{S \mapsto I} (\omega(\mathbf{I}^{-j}_{i}), \varphi(A_{i}^{j}))
\enspace,
\end{align}
where the main difference from Eq. (\ref{eq:s2i_instrument}) is that here in constructing the wrong pairs we sample images from wrong images in the correct instrument category, $\mathbf{I}^{-j}_{i}$, instead of images in wrong instrument categories, $\mathbf{I}_{-i}$. Again, we use the network structure as in Fig.~\ref{fig:model} (a).

\noindent \textbf{I2S Generation} \quad 
We train a single I2S model over the entire dataset so that it can generate sound magnitude spectrograms of different instruments from different musical performance images. In other words, the same model can geneFor example, the model generates a sound spectrogram of drum given an image that has a person playing drum. The generator should not make mistakes on the type of instruments while generating convertible spectrogram to realistic sounds. In this case, we set the training as following: 
\begin{align}
\hat{x}_A &\gets G_{I \mapsto S} (\phi(I_{i}^{j}),z) \nonumber \\
S_f &= D_{I \mapsto S} (\hat{x}_A,\phi(I_{i}^{j})) \nonumber \\
S_r &= D_{I \mapsto S} (A_{i}^{j},\phi(I_{i}^{j})) \nonumber \\
S_w &= D_{I \mapsto S} (\omega(\mathbf{A}_{-i}),\phi(I_{i}^{j}))
\enspace.
\end{align}
Recall that $\hat{x}_A$ is the generated sound spectrogram with size 128x34, and $\phi(I_{i}^{j})$ is the compressed image encoding. We use the image-to-sound network as in Fig.~\ref{fig:model} (b).

\subsection{Sound Encoder Network}
\label{sec:model:audioencode}
\begin{figure}[t]
\centering
\includegraphics[width=\linewidth]{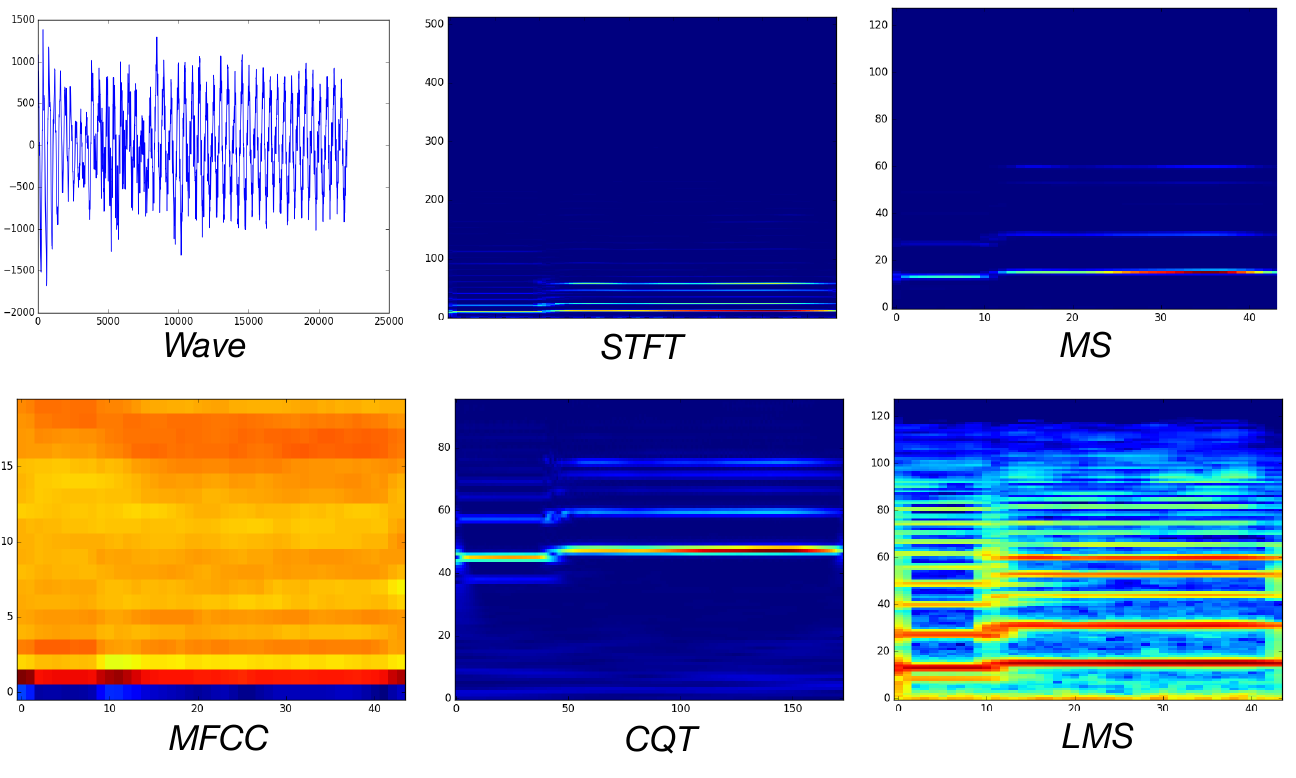}
\caption{Different representations of audio that are fed to the encoder network. The horizontal axis is time and the vertical axis is amplitude (for Wave), frequency (for STFT, MS, CQT, and LMS) or quefrency (for MFCC).}
\label{fig:audio_transform}
\end{figure}

The sound files are sampled at 44,100 Hz. To encode sound, we first transform the raw audio waveform into the time-frequency or time-quefrency domain. We explore a set of representations including the Short-Time Fourier Transform ($STFT$), Constant-Q Transform ($CQT$), Mel-Frequency Cepstral Coefficients ($MFCC$), Mel-Spectrum ($MS$) and Log-amplitude of Mel-Spectrum (LMS). Figure~\ref{fig:audio_transform} shows images of the above-mentioned representations for the same sound. We can see that $LMS$ shows clearer patterns than other representations.
%
%

\begin{table}[!htbp]
\centering
\begin{tabular}{c|c|c|c|c|c}
Accuracy &$MS$ & $LMS$ &CQT & MFCC & STFT \\\hline
3 layers&$62.01\%$& $\textbf{\textit{84.12 \%}}$ & $73.00 \%$ & $80.06 \%$ & $74.05 \%$\\
4 layers&$66.09\%$& $\textbf{\textit{87.44 \%}}$ & $77.78 \%$ & $81.05 \%$ & $75.73 \%$\\
\end{tabular}
\caption{Accuracy of audio classifier. We apply three Conv layers and four Conv layers respectively and it shows the best performance is using four Conv layers.}
\label{tab:t3}
\end{table}

\begin{figure}[!htbp]
\centering
\includegraphics[width=0.8\linewidth]{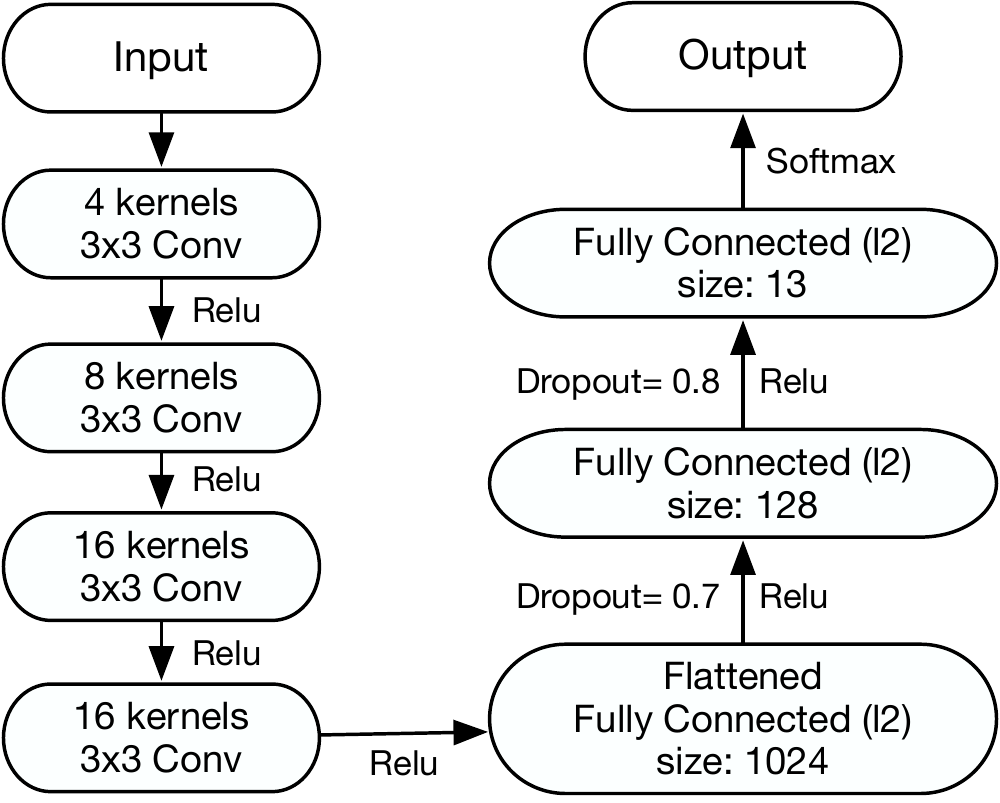}
\caption{Audio classifier trained with instrument category loss.}
\label{fig:audio_classifier}
\end{figure}

We further run a CNN-based classifier on these different representations. We use four convolutional layers and three fully connected layers (see Fig.~\ref{fig:audio_classifier}). In order to prevent overfitting, we add penalties (l2 = 0.015) on layer parameters in fully connected layers, and we apply dropout ($0.7$ and $0.8$ respectively) to the last two layers. The classification accuracies obtained by different representations are shown in Table~\ref{tab:t3}. We can see that $LMS$ shows the highest accuracy. Therefore, we chose $LMS$ over other representations as the input to the audio encoder network. Furthermore, $LMS$ is smaller in size as compared to $STFT$, which saves the running time. Finally, we feed the output of the FC layer (size: 1x128) of CNNs classifier to GAN network as audio feature.

Further merit of $LMS$ is detailed in the experiment section. We thus choose $LMS$ to represent the audio. To calculate $LMS$, a Short-Time Fourier Transform (STFT) with a 2048-point FFT window with a 512-point hop size is first applied to the waveform to get the linear-amplitude linear-frequency spectrogram. Then a mel-filter bank is applied to warp the frequency scale into the mel-scale, and the linear amplitude is converted to the logarithmic scale as well.




\subsection{Image Encoder Network}
\label{sec:model:visualencode}

\begin{figure}[!htbp]
\centering
\includegraphics[width=0.7\linewidth]{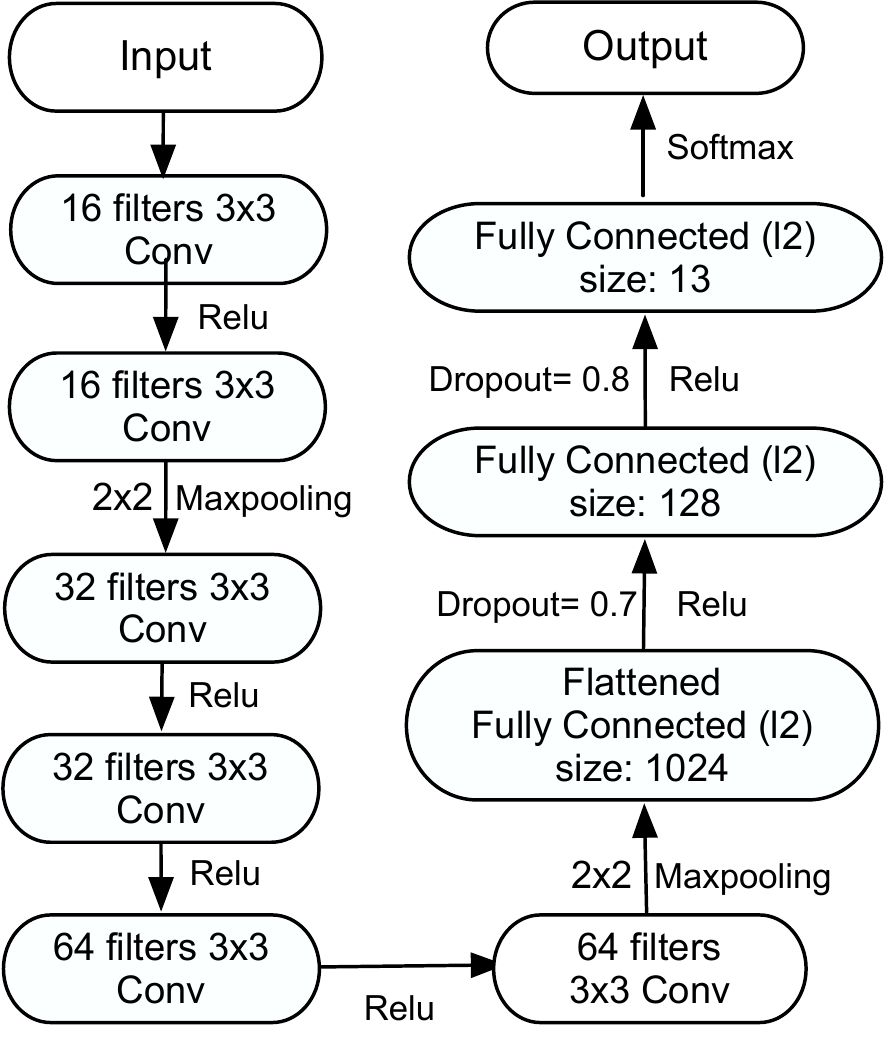}
\caption{Image classifier trained with instrument category loss.}
\label{fig:image_classifier}
\end{figure}

\begin{figure*}[!htbp]
\centering
\includegraphics[width=\linewidth]{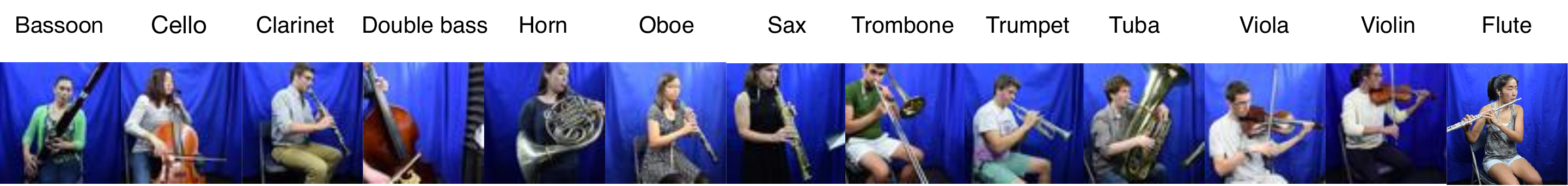}
\caption{Example in the Sub-URMP dataset. Each category contains roughly 6 different complete solo songs.}
\label{fig:newdata}
\end{figure*}

\begin{table*}[!htbp]
\centering
\begin{tabular}{c|c|c|c|c|c|c|c|c|c|c|c|c|c}
Category & Cello & Double Bass &Oboe & Sax & Trumpet & Viola&Bassoon&Clarinet&Horn&Flute&Trombone&Tuba&Violin\\\hline
Training set&$1619$& $448$ & $626$ & $1203$ & $1138$  &$1708$& $138$ & $1308$ & $774$ & $1820$  & $1433$ & $855$ & $1263$  \\
Testing Set&$289$& $245$ & $465$ & $217$ & $285$ &$177$& $260$ & $337$ & $145$ & $327$ & $278$ & $136$ & $341$ \\
\end{tabular}
\caption{Number of image-sound pairs in the Sub-URMP dataset.}
\label{tab:t6}
\end{table*}

For encoding images, we train a CNN with six convolutional layers and three fully connected layers (see Fig.~\ref{fig:image_classifier}). All the convolution kernels are of size 3x3. The last layer is used for classification with softmax loss. This CNN image classifier achieves a high accuracy of more than 95 percent on the testing set. After the network is trained, its last layer is removed, and the feature vector of the second to the last layer having size 128 is used as the image encoding in our GAN network.

\section{Datasets}
\label{sec:data}

To the best of our knowledge, there is no existing dataset that we can directly work on. Therefore, we compose two novel datasets to train and evaluate our models, and they are a Subset of URMP (Sub-URMP) dataset and a ImageNet Image-Sound (INIS) dataset.

Sub-URMP dataset is composed from the original URMP dataset~\cite{LiLiDiARXIV2016}. It contains 13 music instrument categories.
In each category, there are recorded videos of 1 to 5 persons playing different music pieces (see Fig.~\ref{fig:newdata}). We separate videos into 80\% for training and 20\% for testing and ensure that a video will not appear in both training and testing sets. We segment the videos into small chunks with a 0.5 second duration. We use the first frame in each chunk to represent the matching image of the audio. We calculate the loudness ($\Gamma$, unit: dBFS) for all audio chunks using the formula $\Gamma=20*log_{10}(|\psi|/\max(\psi))$, where $\psi$ is the matrix after loading wave file into numpy array. We set a threshold ($\Theta = -45$ dBFS) and delete chunks having $\Gamma \leq \Theta$. Finally, there are a total of $17,555$ sound-image pairs in our composed Sub-URMP dataset. The basic information is shown as Table \ref{tab:t6}. We use this dataset as our main dataset to evaluate models in Sec.~\ref{sec:exp}.

\begin{figure}[t]
\centering
\includegraphics[width=0.9\linewidth]{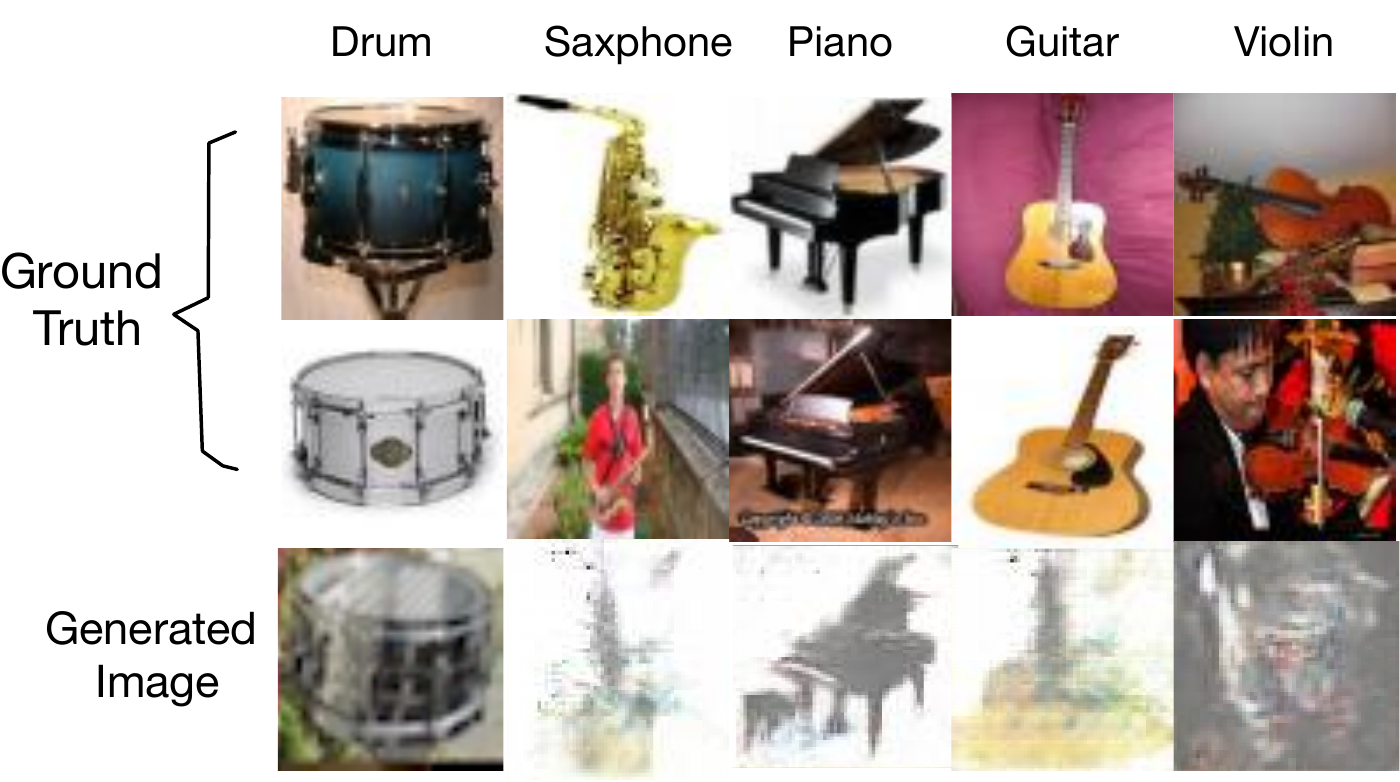}
\caption{Examples from the INIS dataset. Bottom row contains generated images by our S2I-A model .
Due to large variation, images are not as good as those generated in the Sub-URMP dataset.}
\label{fig:olddata}
\end{figure}

All images in the INIS dataset are collected form ImageNet, shown in Fig.~\ref{fig:olddata}. There are five categories, and each contains roughly $1200$ images. In order to eliminate noise, all images are screened manually. Audio files of this dataset come form a total of 77 solo performances downloaded from the Internet, such as a piano performance of the \textit{Moonlight Sonata} and a violin performance of the \textit{Preludio}. We sample $7200$ small audio chunks from all songs with each having 0.5 second duration. We match the audio chunks to the instrument images to create manual sound-image pairs. Table~\ref{tab:t5} shows the statistics of this dataset. 


\begin{table}[t]
\centering
\begin{tabular}{c|c|c|c|c|c}
Category & Piano & Saxphone &Violin & Drum & Guitar \\\hline
Complete songs &$23$& $7$ & $21$ & $7$ & $19$\\
Training set &$766$& $1171$ & $631$ & $1075$ & $818$\\
Testing set &$327$& $500$ & $269$ & $460$ & $349$\\

\end{tabular}
\caption{Distribution of image-sound pairs in INIS dataset}
\label{tab:t5}
\end{table}

\section{Experiments}
\label{sec:exp}

We first introduce our model variations in Sec.~\ref{sec:exp:variations}. Then we present our evaluation on instrument-oriented Sound-to-Image (S2I) generation in Sec.~\ref{sec:exp:s2i_instrument}, pose-oriented S2I generation in Sec.~\ref{sec:exp:s2i_pose} and Image-to-Sound (I2S) generation in Sec.~\ref{sec:exp:i2s}. 

\subsection{Model Variations}
\label{sec:exp:variations}

We have three variations for our sound-to-image network. 

\noindent \textbf{S2I-C network} \quad This is our main sound-to-image network that uses classification-based sound encoding. The model is described in Sec.~\ref{sec:model}.

\noindent \textbf{S2I-N network} \quad This model is a variation of the S2I-C network. It uses the same sound encoding but it is trained without the mismatch $S_w$ information (see Eq.~\ref{eq:s2i_instrument}).  

\noindent \textbf{S2I-A network} \quad This model is a variation of the S2I-C network and differs in that it uses autoencoder-based sound encoding. Here, we use a stacked convolution-deconvolution autoencoder to encode sound. We use four stacks. For the first three stacks, we apply convolution and deconvolution, where the output of convolution is given as input to the next layer in stacks. In the last stack, the input (a 2D array of shape 120x36) is flattened and projected to a vector of size 128 via a fully connected layer. The network is trained to minimize MSE for all stacks in order.

\subsection{Evaluating Instrument-Oriented S2I Generation}
\label{sec:exp:s2i_instrument}

We show qualitative examples in Fig.~\ref{fig:result_training_part} for S2I generation. It can be seen that the quality of the images generated by S2I-C is better than its variations. This is because the classifier is explicitly trained to classify the instruments in sound. Therefore, when this encoding is given as a condition to the generator network, it faces less ambiguity in deciding what to generate. Furthermore, while training the classifier, we observe the classification accuracy, which is a direct measurement of how discriminative the encoding is. This is not true in the case of autoencoder. We know the loss function value, but we do not know if it is a good condition feature in our conditional GANs. 


\subsubsection{\textbf{Human Evaluation}} We have human subjects evaluate our sound-to-image generation. They are given 10 sets of images for each instrument. Each set contains four images; they are generated by S2I-C, S2I-N and S2I-A and a ground-truth image to calibrate the scores. Human subjects are well-informed about the music instrument category of the image sets. However they are not aware the mapping between images to methods. They are asked to score the images on a scale of 0 to 3, where the meaning of each score is given in Table.~\ref{tab:scoremeaning}.

\begin{table}[t]
\centering
\begin{tabular}{c|c|c|c|c|c}
Score & Meaning  \\\hline
3 & Realistic image \& match instrument \\
2 & Realistic image \& mismatch instrument \\
1 & Fair image (player visible, instrument not visible) \\
0 & Unrealistic image
\end{tabular}
\caption{Scoring guideline of human evaluation.}
\label{tab:scoremeaning}
\end{table}

\begin{figure}[t]
\centering
\includegraphics[width=\linewidth]{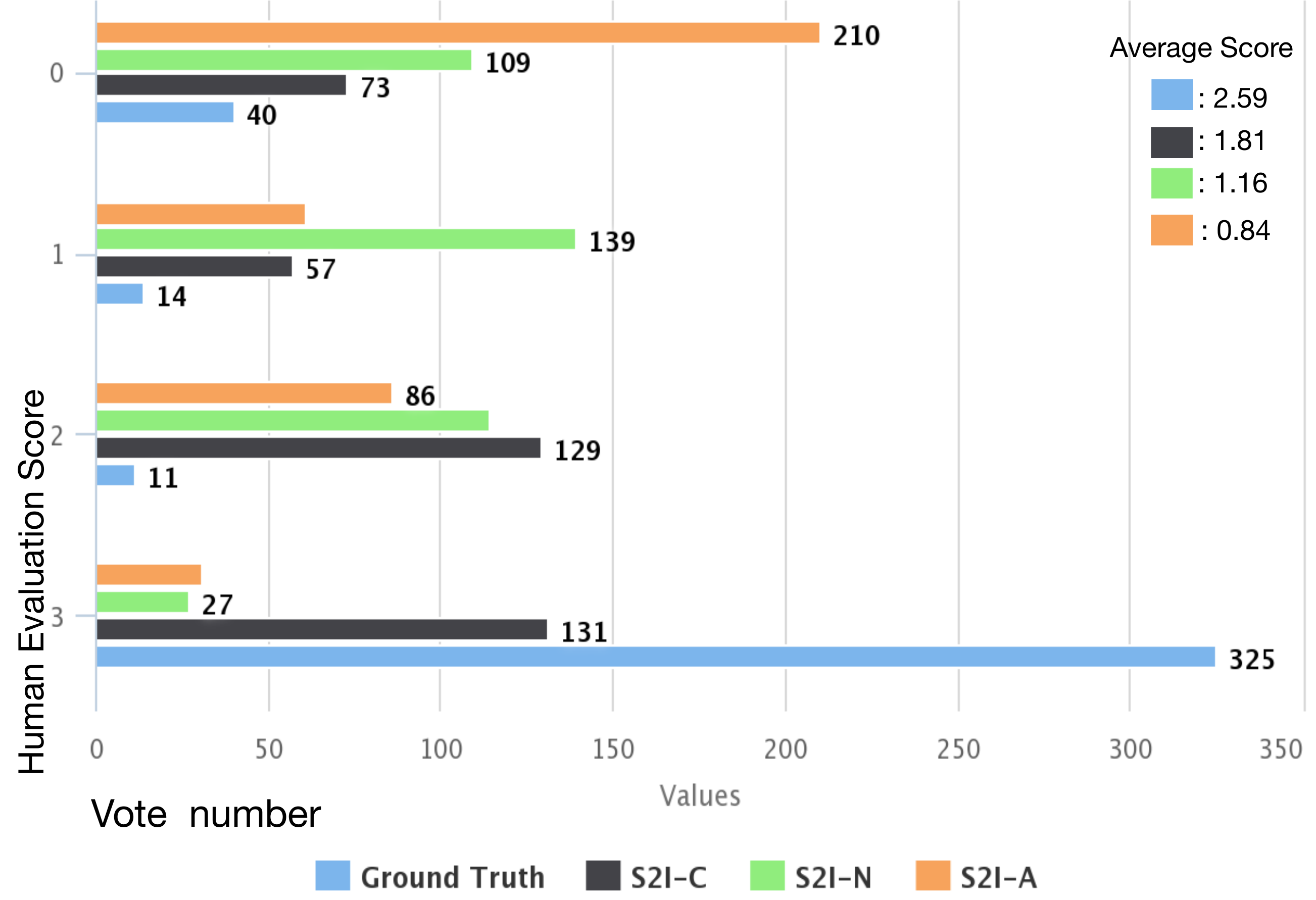}
\caption{Result of human evaluation on generated images. The upper right shows average scores of S2I GANs on human evaluation.}
\label{fig:human_vote}
\end{figure}

Figure~\ref{fig:human_vote} shows the results of human evaluation. More than half of all images generated by S2I-C are considered as realistic by our human subjects, i.e. getting score 2 or 3. One third of them get score 3. This is much higher than S2I-N and S2I-A. In terms of mean score, S2I-C gets 1.81 where the ground-truth gets 2.59 due to small size; all images are evaluated at size 64x64. 

Images from three instruments in particular were rated of very high score among all images generated by S2I-C. Out of 30 Cello images, 18 received highest score of 3, while 25 received scores of 2 or above. Cello images received an average score of 1.9. Out of 30 Flute images, 15 received highest possible score of 3, while 24 received a score of 2 or above. Flute images also received an average score of 2.1.
Out of 30 Double-Bass images, 18 received score 3, while 21 got a score of 2 or more. The average score that Double-Bass images got was 2.02.

\subsubsection{\textbf{Classification Evaluation}}

We use the classifier used for encoding images (see Fig.~\ref{fig:image_classifier}) for evaluating our generated images. When classifying real images, the accuracy of classifier is above $95\%$, thus we decide to use this classifier ($\Gamma$) to verify whether the generated (fake) images are classifiable and they belong to the expected instrument categories. We calculate the accuracies on images generated by S2I-C, S2I-A and S2I-N. Table.~\ref{tab:s2iaccuracies} shows the results. It shows that the accuracy of S2I-A and S2I-N is far worse than the accuracy of S2I-C.  




\begin{figure}
\centering
\includegraphics[width=\linewidth]{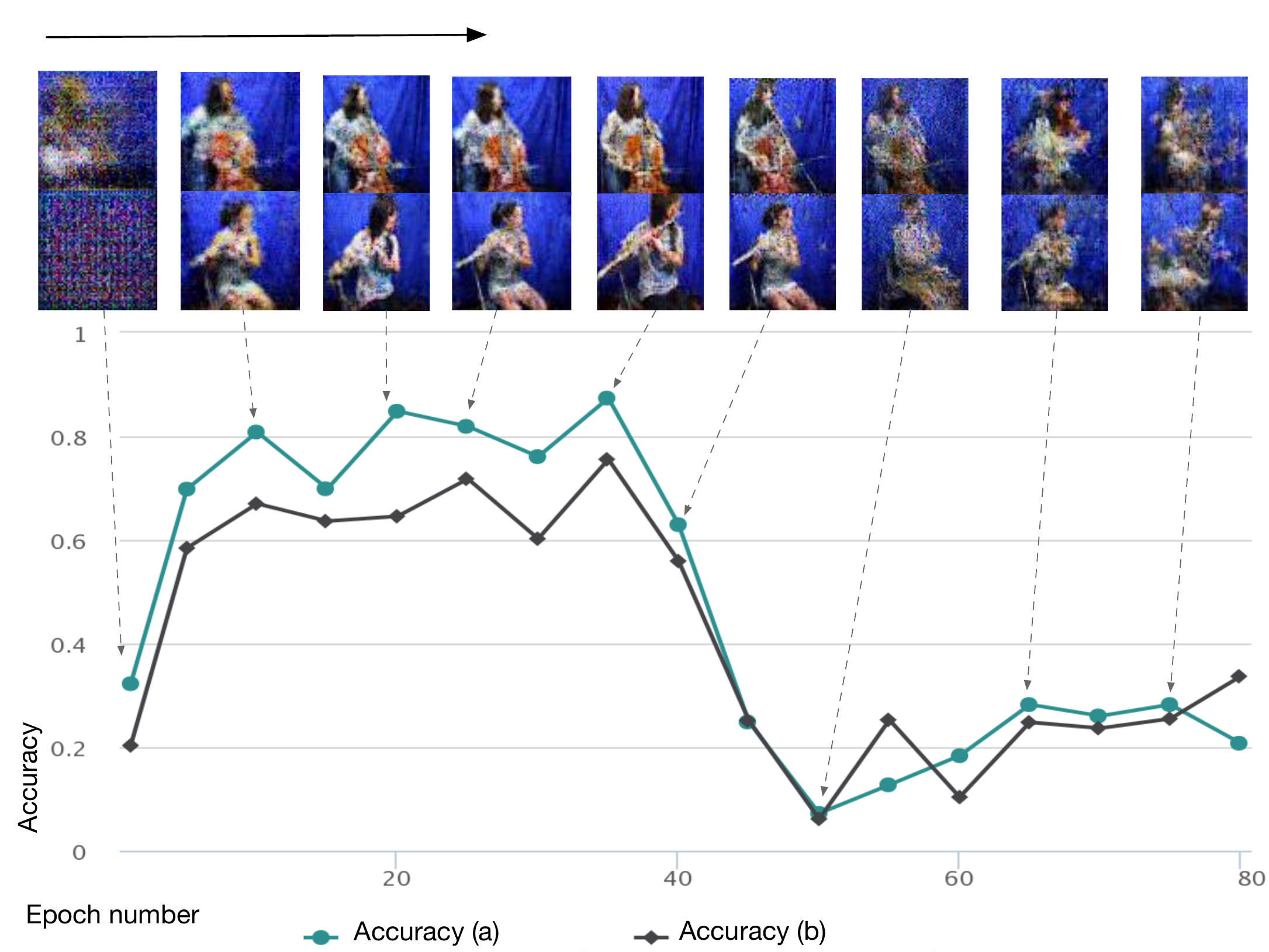}
\caption{Evolution of image quality and classification accuracy on generated images versus number of epochs. Accuracy (a) are the rates that fake images generated in training set of S2I-C are classified into right category by using classifier $\Gamma$. Accuracy (b)  are the rates that fake images generated in testing set of S2I-C are classified into right category by using classifier
$\Gamma$.}
\label{fig:s2iacc}
\end{figure}

\subsubsection{\textbf{Evolution of Classification Accuracy.}} Figure~\ref{fig:s2iacc} shows the classification accuracy on images generated in both training set and testing set. It is plotted for every fifth epoch. The model used for plotting this figure is our main S2I-C network. We visualize generated images for a few key moments in the figure. It shows that the accuracy increases rapidly up till the 35th epoch, and then begins to fall sharply till the 50th epoch, after which it again picks up a little, although the accuracy is still much lower than the peak accuracy. The training and testing accuracies follow nearly the same trend.

One potential reason is that the discriminator loses both classification power and the power to tell fake images apart around epoch 50. Thereafter, it recovers the ability to tell fake images, although not its discriminator power---the slightly higher accuracy is a result of generating the same image with minor variations for all the input audios---thus at least some are classified correctly. This can be seen in the attached images. At epoch 50 we have a totally random looking image, while at 60 we can see that the Cello image looks like the Cello image in the dataset, while the other image, which was supposed to be Flute, looks like the Clarinet image from the dataset. Thus, while the images look like images from the dataset, they are not classified correctly. Hence we get a higher accuracy than bad images, but still not as high as correctly classified, high-quality images.

It is interesting to note that even the fifth epoch has much higher training and testing accuracies than any epoch after 40. What this means is that, even after as few as 5 epochs, not only are the images getting aligned with the expected category, the generated images have enough quality that a classifier can extract distinguishing features from them. This is not true in the case of a random image like the ones in epoch 50.


\begin{table}[t]
\centering
\begin{tabular}{|c|c|c|c|c|c|}
Mode & S2I-C &  S2I-A & S2I-N  \\\hline
Training Set & 87.37\% & 10.63\% & 12.62\%  \\
Testing Set & 75.56\% & 10.95\% & 12.32\% \\

\end{tabular}
\caption{Classifier-based Evaluation Accuracy for Images}
\label{tab:s2iaccuracies}
\end{table}


\subsection{Evaluating Pose-Oriented S2I Generation}
\label{sec:exp:s2i_pose}

\begin{figure}[!htbp]
\centering
\includegraphics[width=\linewidth]{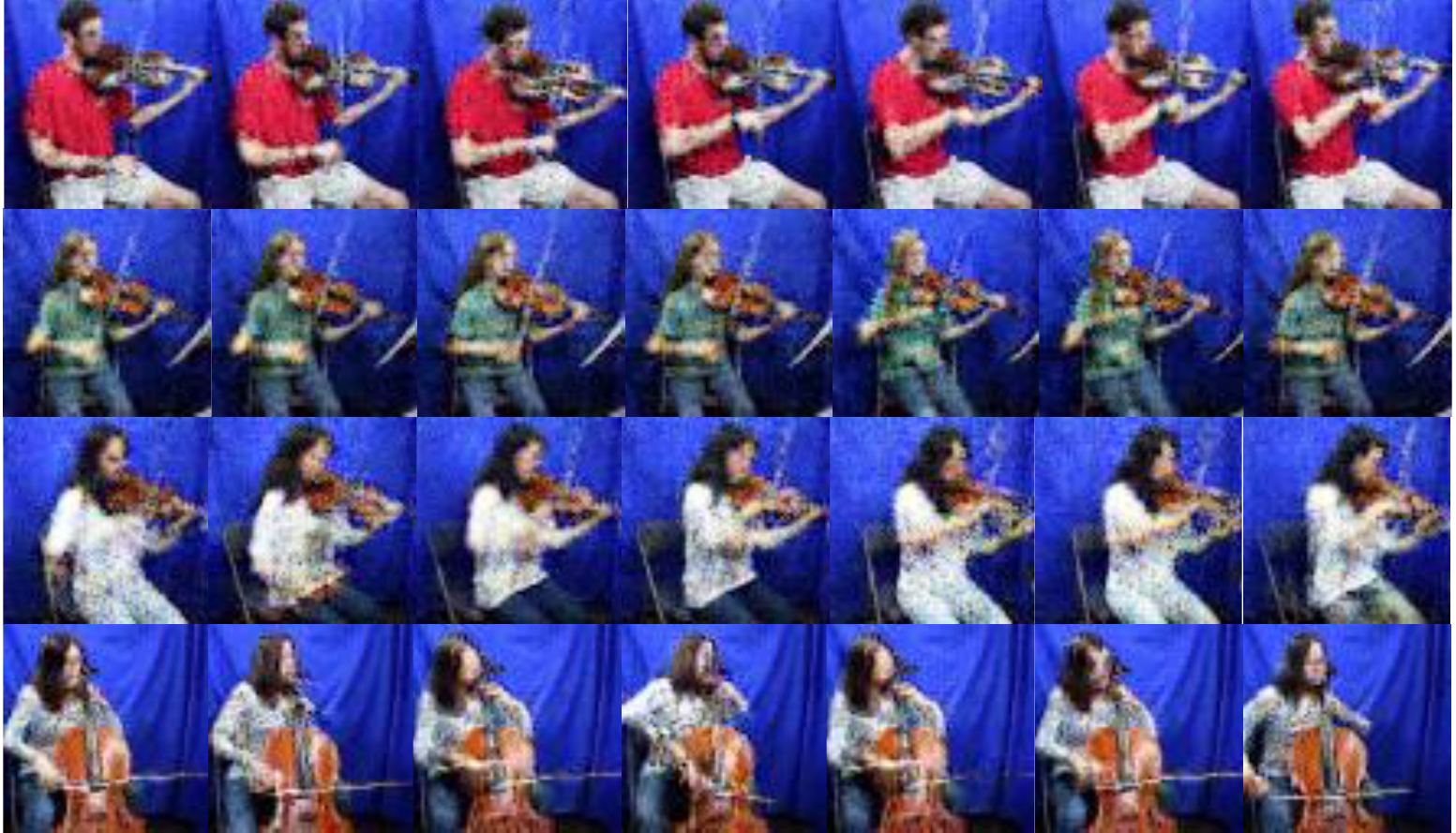}
\caption{Generated pose image. First row shows playing viola, the head position is corresponding to the arm movement. Second and third row are violin, it indicates that one single model can generate multiple persons in different poses. Fourth row is cello, the whole move range is very large. In one single category, songs in training set and testing set are collected from different videos.}
\label{fig:pose}
\end{figure}

The model and the training strategy for our pose-oriented S2I generation is described in Sec.~\ref{sec:model:train}. The results we got were encouraging: various poses can be observed in the generated images (see Fig.~\ref{fig:pose}). Note that for sound encoding, we used the same image classifier as S2I-C. It is trained to classify various instruments, not various poses. With a classifier that is trained to classify music notes, we expect the results to better match the expected poses.

%
%


\subsection{Evaluating I2S Generation}
\label{sec:exp:i2s}

\begin{figure}[t]
\centering
\includegraphics[width=\linewidth]{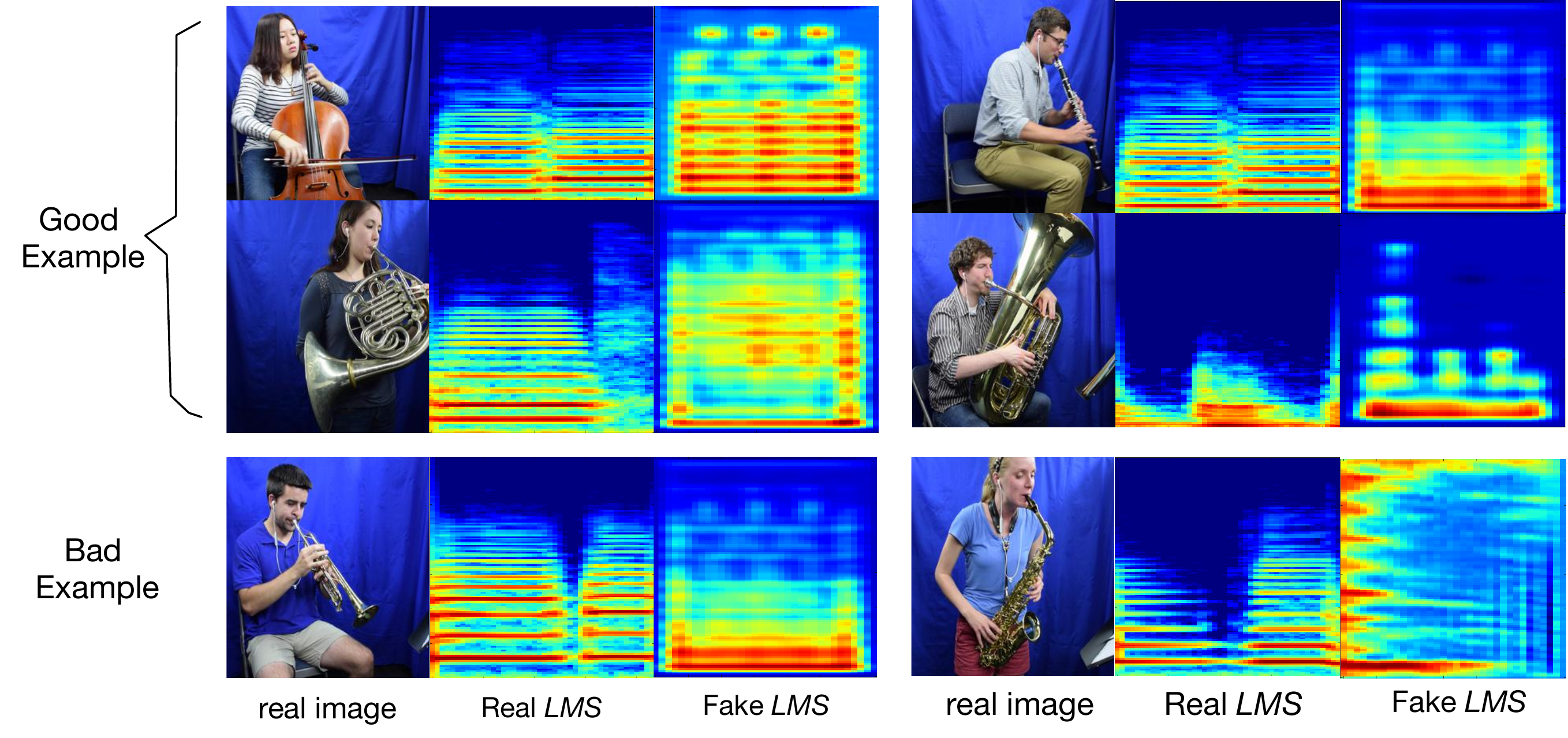}
\caption{Generated sound spectrogram and ground-truth.}

\label{fig:i2s}
\end{figure}

When converting $LMS$ back into waveform files, we will lose high frequency part as the Mel filter filtering is not reversible. Therefore, we conduct evaluation on generated sound spectrograms. We use the sound classifier (see Fig.~\ref{fig:audio_classifier}) which is trained to encode sound for image generation. The reason we use this model is because the model is trained on real $LMS$, and achieves a high accuracy of $80\%$ on testing set of real $LMS$. We achieve $11.17\%$ classification accuracy on the generated $LMS$. One factor that might be affecting the accuracy is that we generate spectrogram of size 128x34 and resize them to 128x44 in classification. Furthermore, Figure~\ref{fig:i2s} shows the generated $LMS$ compared to real $LMS$. We can see, in fake $LMS$, there is less energy in high frequency domain, more energy in low frequency domain, same as real $LMS$. 



\section{Conclusion}
\label{sec:con}

In this paper, we introduce the problem of cross-modal audio-visual generation and make the first attempt to use conditional GANs on intersensory generation. In order to evaluate our models, we compose two novel datasets, i.e., Sub-URMP and INIS. Our experiments demonstrate that our model can, indeed, generate one modality (visual/audio) from the other modality (audio/visual) to a good extent at both instrument-level and pose-level. For example, our model is able to generate pose of a cello player given the note that is being played. 




\noindent \textbf{Limitation and Future Work.} \quad While our I2S model generates LMS, the accuracy is low. Furthermore, it would be worthwhile to hire experts to listen to the ground-truth audio waveform files reconstructed from the generated LMS spectrograms. On the other hand, we are able to generate various poses using our S2I network, but it is hard to quantify how good the generation is. Strengthening the Autoencoder would enable accurate unsupervised generation. The present autoencoder appears to be limited in terms of extracting good representations. It is our future work to explore all these directions. 



\bibliographystyle{ACM-Reference-Format}
\bibliography{vagan} 

\end{document}